\documentclass[pdflatex,sn-basic,Numbered]{sn-jnl}
\usepackage[utf8]{inputenc}
\usepackage{graphicx}
\usepackage{xcolor}
\usepackage{amsmath,amssymb}
\usepackage{booktabs}
\usepackage{multirow}
\usepackage{array}
\usepackage{tabularx}
\usepackage{pifont}
\usepackage[switch]{lineno}
\hypersetup{hidelinks,pdfsubject={}}
\hypersetup{pdfauthor={Hikmat Khan, Metin N. Gurcan, Anil Parwani, Wendy L. Frankel, Wei Chen, Muhammad Khalid Khan Niazi}}
\input{glyphtounicode}
\graphicspath{{./}}
\newcommand{\maybeincludegraphics}[2][]{%
  \IfFileExists{#2}{\includegraphics[#1]{#2}}{%
    \PackageError{MorphContrast}{Missing figure file: #2}{Keep all figures beside the main source.}%
  }%
}
\newcommand{\Lintra}{\mathcal{L}_{\text{intra}}}
\newcommand{\Linter}{\mathcal{L}_{\text{inter}}}
\newcommand{\Lsurv}{\mathcal{L}_{\text{surv}}}
\newcommand{\Ldiv}{\mathcal{L}_{\text{div}}}

\begin{document}
\title{Morphology-aware deep learning predicts colorectal cancer survival from whole-slide images}
\author*[1]{\fnm{Hikmat} \sur{Khan}}
\email{\href{mailto:Hikmat.khan@osumc.edu?cc=Hikmat.khan179@gmail.com}{Hikmat.khan@osumc.edu}}
\author[2]{\fnm{Metin N.} \sur{Gurcan}}
\author[1]{\fnm{Anil} \sur{Parwani}}
\author[1]{\fnm{Wendy L.} \sur{Frankel}}
\author[1]{\fnm{Wei} \sur{Chen}}
\author[1]{\fnm{Muhammad Khalid Khan} \sur{Niazi}}
\affil[1]{\orgdiv{Department of Pathology}, \orgname{The Ohio State University Wexner Medical Center}, \orgaddress{\city{Columbus}, \state{OH}, \country{USA}}}
\affil[2]{\orgdiv{Center for Artificial Intelligence Research}, \orgname{Wake Forest University School of Medicine}, \orgaddress{\city{Winston-Salem}, \state{NC}, \country{USA}}}

\abstract{
\textbf{Background:}
Five-year survival prediction from hematoxylin-and-eosin whole-slide images
could support risk stratification, but the task is weakly
supervised, and the tissue morphologies carrying prognostic information remain
incompletely characterized.

\textbf{Methods:}
We conducted a retrospective secondary analysis using TCGA-COAD/READ
($n{=}562$) for morphology-specific analysis, Alliance/ CALGB~89803
($n{=}424$) for model development, and SurGen ($n{=}218$) for external
evaluation. We propose
MorphContrast, a morphology-aware contrastive multiple-instance learning
framework that uses tissue labels as structural supervision for five-year
survival prediction. MorphContrast combines gated-attention pooling with three auxiliary objectives: patch-level morphology labels define within-patient contrasts, outcome concordance or discordance defines cross-patient contrasts within matched morphologies, and diversity regularization discourages representational collapse. Morphology-matched patches from patients with
discordant outcomes serve as hard negatives, encouraging prognostic
discrimination within comparable histologic contexts.

\textbf{Results:}
Morphology-specific models show weak but distributed prognostic discrimination:
adipose tissue achieves an area under the receiver operating characteristic
curve (AUC) of $0.622$, stroma achieves $0.604$, and both perform comparably to
tumor epithelium ($0.601$--$0.605$). On
Alliance/CALGB~89803, the single model achieves
an AUC of $0.642$, while the post
hoc ensemble reaches a pooled out-of-fold AUC of $0.70$
and a risk-group hazard ratio of $3.79$ ($95\%$
confidence interval: $2.55$--$5.63$). On the external SurGen
colon cancer cohort
($n{=}218$), the single model and
ensemble achieve AUCs of $0.669$ and $0.678$, respectively.

\textbf{Conclusions:}
These findings
support tissue morphology as structural supervision for outcome-aware
representation learning in colorectal cancer prognosis.
}
\keywords{Colorectal cancer, Survival prediction, Computational pathology, Multiple-instance learning, Contrastive learning, Tissue morphology}
\maketitle
\renewcommand{\thelinenumber}{}\linenumbers
\section*{Plain language summary}
Pathologists routinely examine stained tissue sections from colorectal cancers,
and many hospitals now store these sections as digital whole-slide images. We
developed MorphContrast, a computational method that learns from both the tissue
patterns in these images and patients' five-year outcomes. Different tissue
patterns each carried modest prognostic information, suggesting that useful
signals were distributed across the slide rather than confined to tumor cells.
In an archival tissue cohort originating from a clinical trial, the method
distinguished patients with different five-year outcomes, and it showed similar
discrimination in an independent external cohort. The results support the use of tissue-pattern labels to guide
the development of image-based prognostic models, but further prospective and
independent validation is needed before clinical use.

\section{Introduction}
\label{sec:intro}

Colorectal cancer (CRC) is the third most common malignancy and the second
leading cause of cancer-related death worldwide. Its annual burden is projected
to reach $2.2$ million new cases and $1.1$ million deaths by
2030~\cite{rawla2019}, and approximately $154{,}000$ new diagnoses were
anticipated in the United States in 2025~\cite{siegel2025}. Hematoxylin-and-eosin
(H\&E) slides are routinely generated during histopathologic assessment of
surgical resections. Although only an estimated
$15\%$--$20\%$ of US pathology laboratories have adopted digital pathology, this
share is steadily increasing~\cite{kather2026cost,khan2026weakly,datwani2025artificial,akbar2026pathoscribe},
and whole-slide images (WSIs) are already captured wherever slides are digitized.
Predicting five-year survival directly from these WSIs is therefore an
opportunistic use of existing data, offering additional risk stratification at
minimal additional cost.

Two properties make the task difficult. First, supervision is weak: a gigapixel
slide containing thousands of patches carries a single patient-level outcome, so
a model must locate prognostically relevant regions without patch-level
annotation and aggregate them into one prediction~\cite{akbar2026spatially}. Second, it remains unclear
where prognostic information resides. Several established prognostic
features extend beyond malignant epithelium of the main tumor bulk: tumor budding at the invasive front is associated with poor
outcome~\cite{lugli2017itbcc}, the tumor--stroma ratio carries prognostic
information~\cite{gao2021tsr}, and a deep-learning-derived stroma score remains
prognostic after adjustment for standard clinicopathologic
variables~\cite{kather2019}---yet their relative prognostic value is
insufficiently characterized. The distinction bears directly on model design:
signal concentrated in one morphology would justify tissue-restricted analysis,
whereas weak evidence spread across many morphologies would require aggregation
over the complete tissue mixture with unequal weighting.

Existing approaches do not resolve this question. Attention- and
transformer-based methods for multiple-instance learning
(MIL)~\cite{ilse2018abmil,lu2021clam,shao2021transmil,khan2025predicting,tang2024rrtmil}
learn unequal weights over patch representations but derive patch relevance
primarily from the patient outcome. Pathology foundation models provide frozen
features from large-scale, organ-agnostic
pretraining~\cite{chen2024uni,xu2024gigapath}, and morphology-aware CRC models
incorporate tissue information either as an input feature,
as an intermediate representation, or
as an auxiliary
annotation~\cite{kather2019,morphdistill,prism2026}.
However, the approaches considered here do not jointly
organize patch representations by tissue morphology and patient outcome. In
particular, they do not explicitly compare matched morphologies across patients
with concordant and discordant outcomes and therefore do not directly target
prognostic variation \emph{within} a morphology class.

\begin{figure}[!htbp]
\centering
\includegraphics[width=0.9999\linewidth]{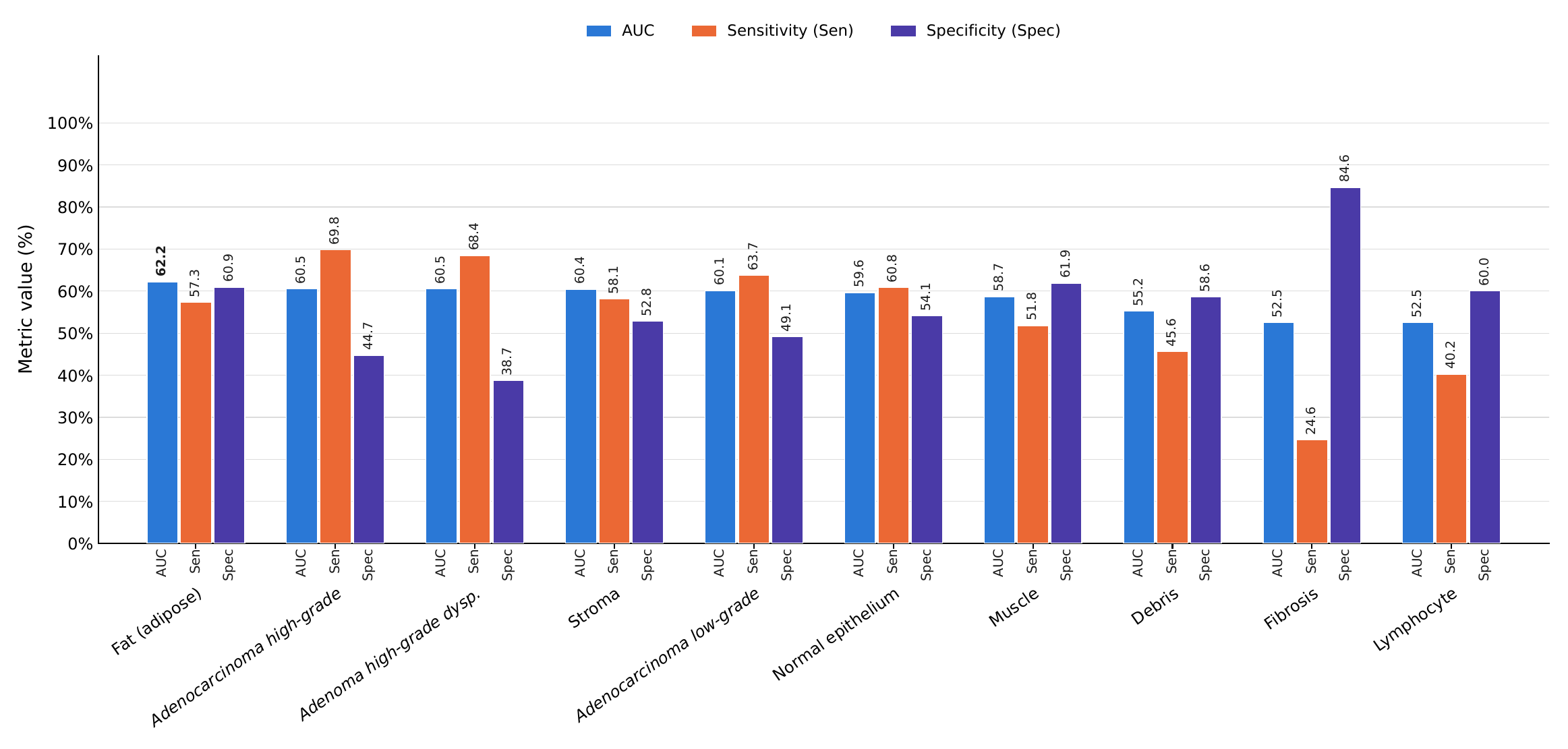}
\caption{\textbf{Top-performing CRC tissue morphologies
show modest prognostic discrimination.} Each bar gives the five-fold
cross-validated AUC of an attention-based multiple-instance learning (ABMIL)
survival model trained on patches of a single morphology class in the
TCGA-COAD/READ cohort ($n{=}562$;
Table~\ref{tab:morph}); the dashed line marks chance ($0.50$).
Only the top-performing morphologies with AUC values above
$0.50$ are displayed. Non-tumor morphologies---adipose tissue, stroma, normal
epithelium, and muscle---score comparably to
tumor-epithelial classes, while the displayed
morphology-specific models remain modest predictors. Because patch abundance,
spatial context, and clinicopathologic correlates are uncontrolled, this
analysis motivates the design of MorphContrast rather than supporting a
biological interpretation.}
\label{fig:teaser}
\end{figure}

We therefore begin with an empirical question: how much prognostic information
does each tissue morphology carry in isolation? The MorphDistill
encoder~\cite{morphdistill} distills morphology-specific knowledge from several
pathology foundation models and assigns CRC morphology
labels to image patches. Using assignments for $17$ tissue morphologies, we
train a separate attention-based MIL survival model
for each morphology in The Cancer Genome Atlas colon and rectal adenocarcinoma
cohorts (TCGA-COAD/READ). Across these morphology-specific models, the prognostic signal is weak and graded rather than concentrated in any single tissue class: several non-tumor morphologies perform comparably to tumor epithelium, yet no individual morphology is strongly predictive(Fig.~\ref{fig:teaser}, Table~\ref{tab:morph}).

We do not interpret this as a direct biological effect,
since spatial context and correlated clinicopathologic factors are uncontrolled. The purpose is methodological:
prognostic evidence is weak, graded, and spread across tumor and non-tumor
morphologies, which motivates using tissue morphology annotations as auxiliary
supervision for patient-level survival prediction rather than only as a means of
distinguishing morphology classes.

This observation motivates the design of \textbf{MorphContrast}, the central contribution of this work. Because
prognostic information is distributed across morphologies, including non-tumor
tissue, MorphContrast processes the complete tissue mixture rather than
restricting analysis to tumor patches. Gated attention
learns the relative contribution of each patch to the patient-level prediction.
An intra-patient contrastive objective ($\Lintra$) organizes patch
representations by morphology within each patient, whereas an inter-patient
objective ($\Linter$) incorporates outcome information: morphology-matched
patches from patients with concordant outcomes form positive pairs, and those
from patients with discordant outcomes form hard negatives. A diversity
regularizer ($\Ldiv$) discourages projected patch representations from
collapsing toward a common direction. These objectives are optimized jointly
with the survival loss, while the MorphDistill patch encoder remains frozen.

Together, these objectives distinguish MorphContrast from conventional attention-based MIL, which learns patch importance
primarily from patient-level outcomes, and from
morphology-aware pipelines in which tissue labels enter as static features or
intermediate predictions~\cite{prism2026}. MorphContrast
instead uses morphology and outcome jointly to define contrastive relationships
that target outcome-associated variation within matched histologic contexts.

We evaluate MorphContrast on Alliance/CALGB~89803 as the
primary cohort and on SurGen~\cite{surgen2025} as an
external cohort. On Alliance/CALGB~89803, the single model reaches
an AUC of $0.642$, and an internally selected post hoc ensemble reaches a pooled
OOF AUC of $0.70$. On SurGen ($n{=}218$), the single model and ensemble achieve AUCs
of $0.669$ and $0.678$, respectively; the ensemble result is similar to
that of FR-MIL, the strongest evaluated baseline ($0.6766$). These results
support morphology-informed representation learning within the evaluated
cohorts but do not establish a clear external performance advantage.

Our contributions can be summarized as follows:
\begin{enumerate}
\itemsep0.15em

\item \textbf{A motivating morphology-specific analysis} that estimates morphology-specific prognostic discrimination for $17$
CRC tissue morphologies using one survival model per class, characterizing the
evidence as weak, graded, and distributed across tumor and non-tumor tissue. It
is offered as methodological motivation rather than as a biological finding.

\item \textbf{MorphContrast}, a morphology-aware contrastive MIL framework in
which tissue morphology labels supervise the structure of prognostic
representations rather than serve only as post hoc annotations. The contrastive
form follows supervised contrastive learning~\cite{khosla2020supcon}; the
contribution lies in jointly optimizing morphology-aware
contrastive objectives with the five-year survival objective.

\item \textbf{Morphology-conditioned outcome contrast},
which aligns morphology-matched patches from patients with
concordant outcomes and treats morphology-matched patches from patients with
discordant outcomes as hard negatives, thereby targeting prognostic variation
within comparable histologic contexts.

\item \textbf{Evaluation on a randomized trial cohort with independent external
assessment}, using Alliance/CALGB~89803 as the primary
cohort and SurGen~\cite{surgen2025} as the external cohort. The single model reaches an AUC of $0.642$, with sensitivity
of $63.53\%$ and specificity of $55.37\%$, on Alliance/CALGB~89803, while the post hoc ensemble reaches an AUC of $0.70$, with sensitivity
and specificity of $68.0\%$ each. On SurGen ($n{=}218$), the single model and
ensemble reach AUCs of $0.669$ and $0.678$, sensitivities of $57.58\%$ and
$65.15\%$, and specificities of $70.39\%$ and $66.45\%$, respectively; the
ensemble AUC is similar to that of FR-MIL, the strongest evaluated baseline
($0.676$).
\end{enumerate}

\section{Related work}
\label{sec:related_work}

MIL represents a WSI as a bag of patches and assigns a single label to the bag \cite{ilse2018abmil}. Gated
attention, CLAM, DSMIL, TransMIL, RRT-MIL, FR-MIL, and PANTHER improve
aggregation through learned weighting, instance clustering, critical-instance
selection, self-attention, regional re-embedding, feature recalibration, and
prototype modeling,
respectively~\cite{ilse2018abmil,lu2021clam,li2021dsmil,shao2021transmil,
tang2024rrtmil,chikontwe2022feature,song2024panther}. Prognostic models such as
those of Nakanishi et al.~\cite{nakanishi} and PROGPATH~\cite{progpath} apply
related weakly supervised principles to recurrence and
survival prediction, respectively. These methods can identify predictive
patches but do not use explicit morphology labels to
structure the prognostic representation~\cite{progpath, nakanishi}.
MorphContrast retains gated-attention pooling and adds morphology- and outcome-aware auxiliary contrastive
objectives together with diversity regularization.

ACMIL reduces excessive attention concentration, whereas MHIM-MIL masks highly
salient instances so that a student learns from harder
instances~\cite{zhang2024acmil,tang2023mhim}. Both
use attention-derived salience to determine which
instances are masked during training. MorphContrast instead defines a hard
negative as a patch with the same morphology as the anchor but from a patient
with an outcome discordant with that of the anchor, so it
changes what the representation encodes rather than which instances the
aggregator receives.

Contrastive learning attracts positive pairs and repels negatives defined by
augmentation or class labels~\cite{chen2020simclr,khosla2020supcon,khan2025morphgen}. Large
pathology foundation models use related objectives to produce transferable
frozen features~\cite{chen2024uni,lu2024conch,xu2024gigapath,
zimmermann2024virchow2,wang2022ctranspath,saillard2024hoptimus,kaiko2024,
kang2023lunit,filiot2024phikon2}. Because
these pretraining objectives do not explicitly use CRC
morphology labels or patient outcomes, they do not constrain the resulting
features to organize tissue identity and patient outcome jointly. MorphContrast
adopts the supervised-contrastive formulation~\cite{khosla2020supcon} but
constructs its pairs from morphology and outcome and optimizes
the corresponding objectives jointly with survival
prediction.

CRC studies have shown that non-tumor morphology can carry prognostic
information~\cite{kather2019}. MorphDistill~\cite{morphdistill} provides
a patch embedding together with an $18$-class morphology
label, and PRISM~\cite{prism2026} fuses morphology-informed and foundation-model
features through cross-feature interaction and gated
attention. These methods use
morphology as an input, an intermediate representation,
or auxiliary supervision. MorphContrast likewise starts
from the patch embeddings and labels provided by the frozen MorphDistill
encoder but uses morphology and outcome jointly to define
representation-learning relationships, targeting prognostic variation within
matched histologic contexts.

Together, the analyses show that morphology labels can provide structural
supervision for outcome-aware representations. The externally assessed
discrimination is comparable to the strongest evaluated baseline, although
prospective validation is required before clinical use.

\begin{figure}[!htbp]
\centering
\includegraphics[width=0.95\linewidth]{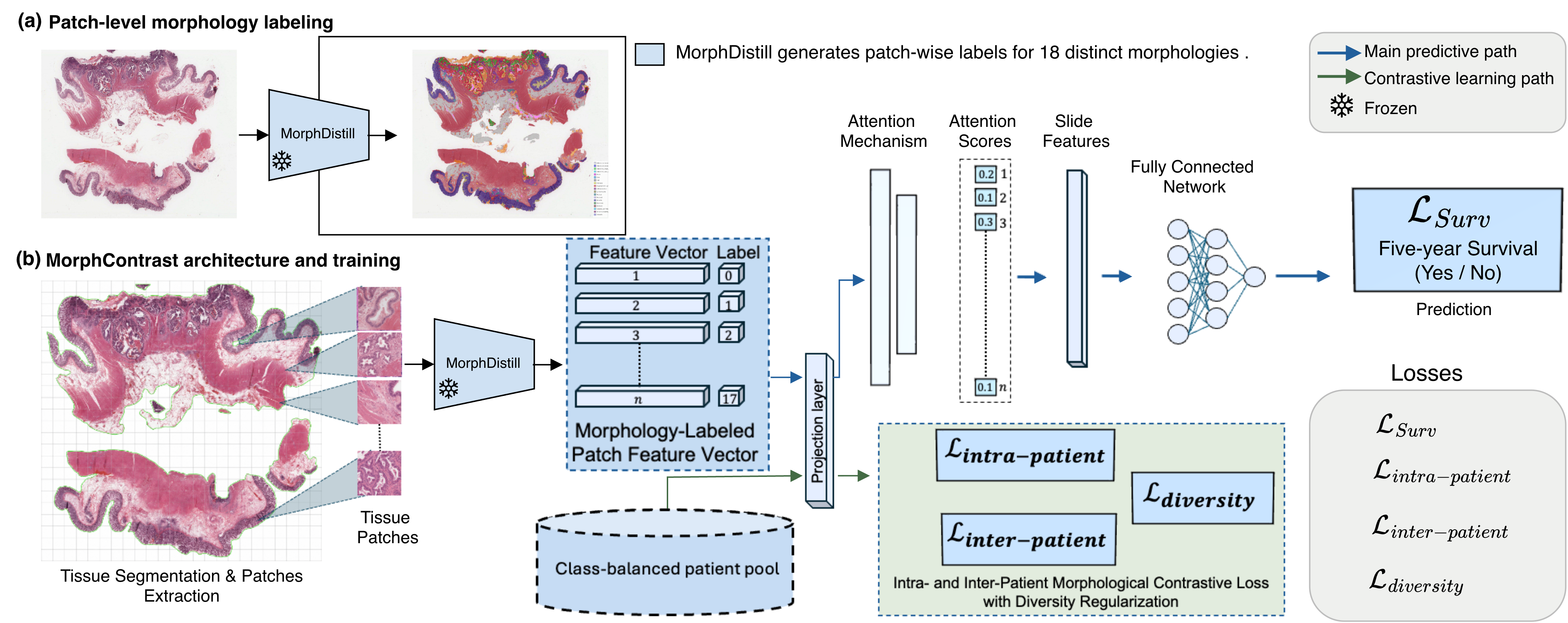}
\caption{\textbf{Overview of MorphContrast.}
\textbf{(a)}~A frozen MorphDistill encoder~\cite{morphdistill} assigns every
patch one of $18$ CRC morphology classes, shown as a color overlay on a
whole-slide image; this step is not trained here. \textbf{(b)}~Tissue is
segmented and tiled, and the same frozen encoder yields a feature vector and a
morphology label $m_{bi}\in\{0,\ldots,17\}$ for each patch. A shared feature
encoder then supplies two pathways. Along the predictive path (blue arrows),
gated attention weights the patches and a fully connected network maps the
pooled representation $\bar{h}_b$ (labeled ``Slide Features'') to the five-year
survival prediction under $\Lsurv$. Along the contrastive path (green arrows), a
projection head is trained with intra-patient morphology contrast ($\Lintra$),
outcome-aware inter-patient contrast ($\Linter$), and diversity regularization
($\Ldiv$); the figure writes these as $\mathcal{L}_{intra\text{-}patient}$,
$\mathcal{L}_{inter\text{-}patient}$, and $\mathcal{L}_{diversity}$.
Contrastive pairs are drawn from the class-balanced
patient pool (Section~\ref{sec:memory}), from
which one outcome-concordant and one outcome-discordant patient bag are sampled
per anchor; losses from both pathways are accumulated across the three bags
before a single backward pass. Snowflakes mark frozen components. Only the
feature encoder, attention pooler, and survival classifier are retained at
inference.}
\label{fig:framework}
\end{figure}

\section{Methods}
\label{sec:method}

\subsection{Overview}
\label{sec:overview}

Figure~\ref{fig:framework} summarizes MorphContrast. A frozen MorphDistill
encoder~\cite{morphdistill} provides a $768$-dimensional feature vector and a
morphology label for each retained patch. A shared feature encoder adapts these
features for two pathways: gated-attention MIL produces a patient-level
five-year survival prediction, while a projection head supports morphology-
and outcome-aware contrastive learning. For each anchor patient, the training
procedure samples one different patient with the same outcome and one with an outcome discordant with that of the anchor from a class-balanced pool constructed exclusively from the training partition
of the current data split. Patches from the anchor and
sampled patients define intra-patient morphology pairs, inter-patient
outcome-aware pairs, and a diversity penalty. All losses are accumulated before
a single backward pass. At inference, the patient pool and projection head are
removed.

\subsection{Architecture}
\label{sec:arch}

Let patient $b$ be represented by $N_b$ patches with precomputed
MorphDistill features given by
$X_b=\{x_{bi}\}_{i=1}^{N_b}$, where $x_{bi}\in\mathbb{R}^{768}$,
with corresponding morphology labels
$m_{bi}\in\{0,\ldots,17\}$ and a patient-level outcome
$y_b\in\{0,1\}$, where $y_b=1$ denotes death within five years and
$y_b=0$ denotes survival for at least five years. The MorphDistill encoder
remains frozen throughout training.

\medskip\noindent\textbf{Shared feature encoder.}\quad
A two-layer task-adaptation network maps each frozen patch feature to an
encoded representation of the same dimensionality:
\begin{equation}
h_{bi}=g_{\theta}(x_{bi})\in\mathbb{R}^{768}.
\label{eq:feature-encoder}
\end{equation}
The encoded patches $h_{bi}$ are shared by the survival and contrastive
pathways, allowing losses from both pathways to shape
the representations used for patient-level prediction
(Fig. \ref{fig:framework}).

\medskip\noindent\textbf{Projection head.}\quad
During training, each encoded patch is mapped to a lower-dimensional
contrastive space:
\begin{equation}
z_{bi}
=
\frac{f_{\phi}(h_{bi})}
{\lVert f_{\phi}(h_{bi})\rVert_2},
\qquad
z_{bi}\in\mathbb{R}^{d_p},
\qquad
\lVert z_{bi}\rVert_2=1,
\qquad
d_p\in\{128,256\}.
\label{eq:projection}
\end{equation}
The projection head $f_{\phi}$ contains two linear layers separated by a ReLU
activation. For the pairwise objectives below, patches
are indexed across the sampled patients by $i$, with $z_i$, $b_i$, and $m_i$
denoting the projected representation, patient identity, and morphology label,
respectively. Because each projected representation has unit $L_2$ norm, the
inner product of two projected patches equals their cosine similarity. We
define the temperature-scaled similarity as
\begin{equation}
s_{ij}
=
\frac{z_i^{\top}z_j}{\tau}.
\label{eq:similarity}
\end{equation}
Here, $\tau>0$ controls the concentration of the contrastive distribution. We
restrict $\tau\in(0,1]$ and set $\tau=0.1$ in all experiments. The projection
head is used only during training and is removed at inference.

\medskip\noindent\textbf{Gated-attention aggregation.}\quad
The survival pathway applies gated attention to the encoded patches. Following
ABMIL~\cite{ilse2018abmil}, the attention weight assigned to patch $i$ is
\begin{equation}
a_{bi}
=
\operatorname{softmax}_{i}
\left(
w^{\top}
\left[
\tanh(W_vh_{bi})
\odot
\sigma(W_uh_{bi})
\right]
\right).
\label{eq:attn}
\end{equation}
Here, $W_v,W_u\in\mathbb{R}^{d_a\times768}$ and
$w\in\mathbb{R}^{d_a}$ are trainable parameters, $d_a$ is the attention hidden
dimension, $\odot$ denotes elementwise multiplication, and $\sigma$ is the
logistic sigmoid. The softmax is computed over the $N_b$ patches of patient
$b$, such that $\sum_{i=1}^{N_b}a_{bi}=1$.
The resulting patient-level representation is
\begin{equation}
\bar{h}_b
=
\sum_{i=1}^{N_b}a_{bi}h_{bi}.
\label{eq:pool}
\end{equation}
This aggregation allows individual patches and tissue morphologies to
contribute unequally to the patient representation.

\medskip\noindent\textbf{Survival classifier.}\quad
A classifier maps the patient representation $\bar{h}_b$ to a scalar survival
logit:
\begin{equation}
\eta_b=q_{\psi}(\bar{h}_b),
\qquad
\hat{p}_b=\sigma(\eta_b),
\label{eq:prediction}
\end{equation}
where $\hat{p}_b$ is the predicted probability of death within five years.

\subsection{Class-Balanced Patient Pool}
\label{sec:memory}

Inter-patient contrast requires patients from both outcome groups. For each
data split, inter-patient contrast uses a class-balanced patient pool
constructed exclusively from the training partition, preventing patient-level
information leakage. The training pool contains equal numbers of patients who
survived for at least five years and patients who died within five years. For
each anchor patient, one different outcome-concordant patient and one
outcome-discordant patient are sampled from this training-only pool and
processed by the shared feature encoder. Up to $512$ morphology-balanced
patches are retained per patient to limit memory use and reduce dominance by
highly abundant tissue compartments. Each retained patch is associated with
its projected representation, patient identity, morphology label, and patient
outcome:
\begin{equation}
(z_i,b_i,m_i,y_{b_i}).
\label{eq:patch-metadata}
\end{equation}
These attributes define the patch-level positive and negative sets used by the
contrastive objectives.

\subsection{Patch-Level Contrastive Objectives}
\label{sec:contrastive}

For an anchor patch $i$ with positive set $P(i)$ and negative set $N(i)$, we
use the sum-over-positives supervised contrastive objective
\begin{equation}
\ell(i;P,N)
=
-\log
\frac{
\displaystyle\sum_{p\in P(i)}
\exp(z_i^{\top}z_p/\tau)
}{
\displaystyle\sum_{k\in P(i)\cup N(i)}
\exp(z_i^{\top}z_k/\tau)
}.
\label{eq:supcon}
\end{equation}
The anchor itself is excluded from both sets. Anchors without at least one
valid positive and one valid negative are omitted from the corresponding loss.

\medskip\noindent\textbf{Intra-patient morphology contrast.}\quad
The intra-patient objective is designed to organize patch
representations by tissue morphology within each patient. For anchor patch
$i$, positives are patches from the same patient with the same morphology,
$P_a(i)=\{j\neq i:b_j=b_i,\;m_j=m_i\}$,
whereas negatives are patches from the same patient with a different
morphology,
$N_a(i)=\{j:b_j=b_i,\;m_j\neq m_i\}$.
The intra-patient loss is
\begin{equation}
\Lintra
=
\frac{1}{|\mathcal{A}_a|}
\sum_{i\in\mathcal{A}_a}
\ell(i;P_a,N_a),
\label{eq:intra}
\end{equation}
where $\mathcal{A}_a$ contains anchors with at least one same-morphology
positive and one different-morphology negative.

\medskip\noindent\textbf{Inter-patient outcome-aware contrast.}\quad
The inter-patient objective uses patient outcome to
define contrastive relationships within each morphology class. For anchor
patch $i$, positives are morphology-matched patches from a different patient
with the same outcome,
$P_e(i)=\{j:b_j\neq b_i,\;m_j=m_i,\;y_{b_j}=y_{b_i}\}$.
Morphology-matched patches from patients with an outcome
discordant with that of the anchor patient form hard negatives,
$N_e^{\mathrm{hard}}(i)=\{j:b_j\neq b_i,\;m_j=m_i,\;y_{b_j}\neq y_{b_i}\}$.
Patches belonging to different morphologies provide standard negatives,
$N_e^{\mathrm{std}}(i)=\{j\neq i:m_j\neq m_i\}$.
The complete negative set is $N_e(i)=N_e^{\mathrm{hard}}(i)\cup
N_e^{\mathrm{std}}(i)$, and the inter-patient loss is
\begin{equation}
\Linter
=
\frac{1}{|\mathcal{A}_e|}
\sum_{i\in\mathcal{A}_e}
\ell(i;P_e,N_e),
\label{eq:inter}
\end{equation}
where $\mathcal{A}_e$ denotes the set of valid inter-patient anchors.
Same-patient, same-morphology patches are excluded from this objective. The
morphology-matched hard negatives encourage the representation to capture
outcome-associated variation within comparable histologic contexts rather
than tissue identity alone.

\medskip\noindent\textbf{Diversity regularization.}\quad
A diversity term discourages the projected patches of a patient from
collapsing toward a common direction~\cite{belagali2026ticon}. For patient $b$
with $K_b$ sampled patches,
\begin{equation}
\ell_b^{\mathrm{div}}
=
\frac{1}{K_b(K_b-1)}
\sum_{i\neq j}
z_i^{\top}z_j,
\qquad
\Ldiv
=
\frac{1}{B}
\sum_{b=1}^{B}
\ell_b^{\mathrm{div}}.
\label{eq:diversity}
\end{equation}
Here, the inner sum is over ordered pairs of distinct
patches from patient $b$, and $B$ is the number of patients contributing to the
diversity term. Minimizing this term encourages
dispersion of a patient's projected patch representations rather than
enforcing exact pairwise orthogonality.

\subsection{Survival Objective and Joint Optimization}
\label{sec:joint}

The patient-level survival loss is computed after gated-attention pooling. For
binary cross-entropy,
\begin{equation}
\Lsurv
=
-\frac{1}{B}
\sum_{b=1}^{B}
\left[
y_b\log\hat{p}_b
+
(1-y_b)\log(1-\hat{p}_b)
\right].
\label{eq:surv}
\end{equation}
Weighted cross-entropy and focal-loss variants are additionally considered
during the ensemble search. The complete training objective is
\begin{equation}
\mathcal{L}
=
\lambda_s\Lsurv
+
\lambda_a\Lintra
+
\lambda_e\Linter
+
\lambda_d\Ldiv.
\label{eq:total}
\end{equation}
The survival-loss weight is fixed at $\lambda_s=1.0$. The auxiliary-loss
weights satisfy $\lambda_a,\lambda_e,\lambda_d\in[0,1]$.
For the model configuration, we use
$\lambda_a=0.25$,
$\lambda_e=0.1$, and
$\lambda_d=0.25$. For each update, $\Lsurv$ is computed from the anchor
prediction, while the auxiliary losses use patches from the anchor and the two
sampled patients. A single backward pass updates the shared feature encoder
through all four loss terms; $\Lsurv$ also updates the
attention network and classifier, whereas the auxiliary losses update the
projection head. Thus, patch-level contrastive supervision reshapes the
features later pooled for survival prediction. At inference, only the
precomputed MorphDistill features, shared feature
encoder, gated-attention pooler, and survival classifier are used.

\subsection{Study design and cohorts}
\label{sec:setup}
\label{sec:cohorts}

This study is a retrospective secondary analysis of digitized archival
histology and pre-existing outcome data. The three cohorts serve distinct roles.
TCGA-COAD/READ is used for the
morphology-specific analysis that empirically characterizes differences in
prognostic value across tissue morphologies and motivates the proposed method
(Section~\ref{sec:results-morph}); Alliance/CALGB~89803 serves as the primary
development cohort and provides the basis for the main comparisons in
Sections~\ref{sec:results-compare}--\ref{sec:results-foundation}; and SurGen
serves as the independent external cohort. All cohorts
use nominal $224\times224$ patches at $20\times$, with patches containing less
than $25\%$ tissue discarded. Tissue regions are segmented
using TRIDENT, and patches are extracted without overlap. The following
paragraphs briefly describe each cohort and its role in the study.

\subsubsection{Primary cohort and trial design}
The primary cohort derives from Alliance/CALGB~89803, a randomized phase~III
trial comparing adjuvant weekly bolus irinotecan with fluorouracil and
leucovorin (IFL) and fluorouracil and leucovorin (FL)
alone in $1{,}264$
patients with completely resected stage~III colon
cancer~\cite{saltz2007,NCT00003835}. CALGB subsequently
became part of the Alliance for Clinical Trials in Oncology; we therefore refer
to the cohort throughout as Alliance/CALGB~89803. Eligible patients had received
no prior chemotherapy or radiotherapy, had performance status $0$--$2$, and
began treatment within $21$--$56$ days of resection, with randomization
stratified by nodal involvement ($1$--$3$ versus $\geq4$ nodes), histological
grade, and preoperative carcinoembryonic antigen. 

\subsubsection{Digitization and quality assurance}
Slides were digitized at $40\times$ ($0.25\,\mu\mathrm{m}$ per pixel) on an
Aperio scanner (Leica Biosystems, Nussloch, Germany) at the participating
academic medical center under an approved institutional review board
protocol~\cite{prism2026}. A trained pathologist, blinded to clinical
information and outcome, reviewed each slide to confirm the presence of tumor;
more than $98\%$ of cases contributed a single representative tumor slide.
Quality control assessed tissue detection and coverage, color fidelity, focus,
and scanning artifacts, and slides with inadequate staining were removed. This study was reviewed and approved by an institutional review board.

\subsubsection{Cohort construction and endpoint}
Cases were excluded for absence of tumor, lymph-node-only specimens,
non-disease-related death within five years, and loss to follow-up before five
years. Patients censored before five years were \emph{excluded rather than
censored}, because the endpoint of interest is cancer-specific rather than
all-cause mortality---a distinction that carries weight in a disease with a
median age at diagnosis of $66$ years and substantial competing
comorbidity~\cite{Rogers2021EarlyOnsetCRC}. This exclusion
produces a complete-case binary endpoint and restricts
the analysis to patients with known five-year outcomes. Mucinous
adenocarcinomas were also excluded, as in prior analyses of this
cohort~\cite{prism2026}, because their architecture precludes reliable
assessment of tumor budding; following the World Health Organization definition
of mucinous carcinoma as a tumor in which extracellular mucin constitutes at
least $50\%$ of tumor volume, $50$ such cases were removed. The resulting cohort
comprises $424$ patients and $424$ WSIs; among the
patients, $103$ died within five years and $321$
survived for at least five years. All tumors arose in the cecum, ascending colon,
hepatic flexure, transverse colon, splenic flexure, descending colon, or sigmoid
colon. Supplementary Table~1 summarizes the clinicopathologic
characteristics.

\subsubsection{Discovery cohort: TCGA-COAD/READ}
The colon and rectal adenocarcinoma collections of The Cancer Genome
Atlas~\cite{TCGA2012Colorectal,GDCDataPortal} provide $562$ patients with
diagnostic formalin-fixed, paraffin-embedded (FFPE) slides
and follow-up and are used exclusively for the morphology-specific empirical
analysis described in Section~\ref{sec:results-morph}.
The morphology-specific models are evaluated using
five-fold cross-validation with patient-level splits. The TCGA cohort differs
from Alliance in scanning protocol and scanner vendor, patient population, and
stage distribution; unlike the Alliance cohort, mucinous cases were retained.
Slides are tiled using the shared protocol of $224\times224$ patches at
$20\times$. Because the TCGA slides were scanned at a
base resolution close to that of Alliance
($\approx0.24\,\mu\mathrm{m}$ per pixel versus $0.25$),
the resulting patches cover a comparable physical field.

\subsubsection{External cohort: SurGen}
\label{sec:surgen-data}
SurGen~\cite{surgen2025} is a public colorectal digital pathology resource
comprising $1{,}020$ H\&E WSIs from $843$ cases collected through NHS Lothian.
We use the survival-annotated SR386 subset, which contains $427$ primary
colorectal cases with five-year outcomes available for $426$;
the external evaluation is restricted to $218$ colon
cancer cases from this subset to align the disease site with that of the
colon-only Alliance cohort. The source subset includes both colon and rectal
cancers across a broader stage distribution than Alliance. All slides were digitized on a ZEISS Axio Scan.Z1 at $40\times$ and
$0.1112\,\mu\mathrm{m}$ per pixel, which differs substantially from the Alliance
acquisition protocol. These differences in scanner
resolution and case mix introduce potential distribution shift between the
development and external cohorts.

\subsection{Comparison Methods}
\label{sec:baselines}

For the Alliance cohort, we compare MorphContrast with four groups of
methods: \emph{(i) Attention- and transformer-based MIL:}
ABMIL~\cite{ilse2018abmil}, CLAM~\cite{lu2021clam},
TransMIL~\cite{shao2021transmil}, and RRT-MIL~\cite{tang2024rrtmil}.
\emph{(ii) Task-specific CRC prognosis:} the
CRC recurrence-prediction model of Nakanishi et
al.~\cite{nakanishi}. \emph{(iii) Morphology-aware CRC prognosis:} MorphDistill
with its standard survival head~\cite{morphdistill} and
PRISM~\cite{prism2026}. \emph{(iv) Frozen pathology
foundation encoders:} UNI~v1/v2~\cite{chen2024uni},
Virchow2~\cite{zimmermann2024virchow2}, GigaPath~\cite{xu2024gigapath},
CONCH~v1.5~\cite{lu2024conch}, CTransPath~\cite{wang2022ctranspath},
H-Optimus-0~\cite{saillard2024hoptimus}, Phikon~v2~\cite{filiot2024phikon2},
Kaiko ViT-B/8~\cite{kaiko2024}, and Lunit ViT-S/8~\cite{kang2023lunit}.
Each frozen encoder is coupled with the same gated-attention survival head to
hold the aggregation architecture constant when comparing
patch representations. Methods in groups (i)--(iii) use their published
configurations and default decision thresholds.

On the external SurGen cohort, we compare MorphContrast with
three MIL baselines: TransMIL~\cite{shao2021transmil}
, FR-MIL~\cite{chikontwe2022feature},
and MHIM-MIL~\cite{tang2023mhim}. These methods represent
transformer aggregation, feature recalibration, and masked hard-instance mining,
respectively. All models reported for SurGen are trained on
Alliance/CALGB~89803 and applied in evaluation mode without additional training;
preprocessing parameters, ensemble membership, and decision thresholds are
fixed before external evaluation. Because the baseline suites differ between
Alliance and SurGen, comparisons are made only within each cohort.

\subsection{Statistical analyses}
\label{sec:metrics}

We evaluate discrimination using the area under the receiver operating
characteristic curve (AUC) and Harrell's concordance index
(C-index)~\cite{Harrell1996PrognosticModels}. We additionally report balanced
accuracy, sensitivity, and specificity at each model's operating threshold,
with death within five years defined as the positive class. Sensitivity,
specificity, and balanced accuracy are computed from true positives (TP), true
negatives (TN), false positives (FP), and false negatives (FN) as
\begin{equation}
\begin{aligned}
\mathrm{Sen} &= \frac{TP}{TP+FN},\\
\mathrm{Spec} &= \frac{TN}{TN+FP},\\
\mathrm{BalAcc} &= \frac{1}{2}\left(\mathrm{Sen}+\mathrm{Spec}\right).
\end{aligned}
\end{equation}

Risk stratification is assessed using the log-rank test and a univariate Cox
proportional-hazards model, with hazard ratios and $95\%$ confidence
intervals~\cite{prism2026}. For Alliance, baselines and
the single MorphContrast model are reported as
mean $\pm$ standard deviation across folds, whereas the ensemble uses pooled
out-of-fold predictions with $1{,}000$-sample bootstrap $95\%$ confidence
intervals. For SurGen, all methods are reported as point
estimates with bootstrap $95\%$ confidence intervals for the independent cohort
of $218$ cases.

\subsection{Implementation and Training Details}
\label{sec:training}

Alliance model development uses five-fold
cross-validation with patient-level splits. For each evaluation split, patients
are allocated in proportions of $70\%$, $15\%$, and $15\%$ to the training,
validation, and test sets, respectively. The validation split is used for
hyperparameter selection and early stopping, and the corresponding held-out test
split is used only to report performance. Models are trained using the AdamW optimizer~\cite{loshchilov2019adamw} for up
to $150$ epochs with early stopping when the validation
loss does not improve for $15$ consecutive epochs. The learning rate follows a
cosine annealing schedule~\cite{loshchilov2017sgdr}. The MorphDistill patch
encoder remains frozen throughout training; only the task-specific feature
encoder, projection head, gated-attention module, and survival classifier are
optimized. Supplementary Table~2 summarizes the hyperparameter search space and
fixed training settings. The initial search over the contrastive-loss weights
selected $\lambda_a=0.25$ and $\lambda_e=0.1$, which are fixed in subsequent
experiments. The MorphContrast uses
binary cross-entropy with $\lambda_s=1.0$ and the default objective weights
described in Section~\ref{sec:joint}. Weighted cross-entropy and focal loss are
additionally evaluated among candidate configurations considered for the post
hoc ensemble.

\subsection{Post Hoc Ensemble Selection from Out-of-Fold Predictions}
\label{sec:ensemble}

Each hyperparameter configuration generates one out-of-fold (OOF) risk score
per patient under five-fold cross-validation. Candidate configurations are
retained only if they achieve $\mathrm{AUC}\geq0.50$, sensitivity
$\geq0.30$, and specificity $\geq0.30$. Ensemble selection varies the number of
members $k\in\{3,5,7,10\}$, survival-loss variant, member-selection strategy,
score-aggregation method, and decision threshold, yielding $50$ candidate
ensembles (Supplementary Table~2). Candidates are ranked using
\begin{equation}
\mathcal{J}_{\mathrm{ens}}
=
\mathrm{AUC}
+
\mathrm{BalAcc}
-
\frac{1}{2}
\left|
\mathrm{Sen}-\mathrm{Spec}
\right|,
\label{eq:ensemble-objective}
\end{equation}
which favors discrimination and balanced classification while penalizing large
differences between sensitivity and specificity. The selected ensemble contains
three Pareto-selected members. Their OOF risk scores are converted to percentile
ranks and aggregated to obtain a single patient-level ensemble score, with a
final decision threshold of $t^{\star}=0.573$. We refer to the single model
as the MorphContrast model and to the ensemble as the post hoc
MorphContrast ensemble.

\section{Results}
\label{sec:results}

\subsection{Prognostic Discrimination Across Individual Tissue Morphologies}
\label{sec:results-morph}

To quantify the prognostic discrimination associated with individual tissue
morphologies, we trained a separate ABMIL survival model for each morphology
class in TCGA-COAD/READ. Among the $17$ non-background classes, four
(low-grade adenoma dysplasia, mucus, necrosis, and sessile serrated lesion)
produced constant predictions, while inflammation and hyperplastic polyp yielded
AUCs of $0.50$. Of the remaining $11$ morphologies, ten achieved AUCs above
$0.50$ and eight exceeded $0.55$; the vessels class was the only class below chance
(AUC $0.466$) (Table~\ref{tab:morph}, Fig.~\ref{fig:teaser}). Adipose tissue achieved the highest AUC ($0.622$), followed by high-grade
adenocarcinoma ($0.605$), high-grade adenoma dysplasia ($0.605$), stroma
($0.604$), low-grade adenocarcinoma ($0.601$), and normal epithelium ($0.596$).
The highest-ranking classes therefore included both tumor-associated and
non-tumor compartments, and no individual morphology exceeded an AUC of
$0.622$.

\begin{table}[!htbp]
\centering
\renewcommand{\baselinestretch}{1}\footnotesize
\footnotesize
\caption{\textbf{Morphology-specific five-year survival prediction.}
Five-fold cross-validated ABMIL models trained independently using patches from
a single morphology class in TCGA-COAD/READ.}
\label{tab:morph}
\newcommand{\TablePoint}[1]{{\fontsize{9}{11}\selectfont\ensuremath{#1}}}
\newcommand{\TableCI}[1]{{\fontsize{6.5}{8}\selectfont\ensuremath{#1}}}
\fontsize{8.5}{10.5}\selectfont
\renewcommand{\arraystretch}{1.08}
\begin{tabularx}{\linewidth}{@{}>{\raggedright\arraybackslash}Xccc@{}}
\toprule
\textbf{Morphology} & \textbf{AUC} & \textbf{Sen} & \textbf{Spec} \\
\midrule
Fat (adipose)                        & \TablePoint{\mathbf{0.622}} & \TablePoint{0.573} & \TablePoint{0.609} \\
Adenocarcinoma high-grade            & \TablePoint{\underline{0.605}} & \TablePoint{0.698} & \TablePoint{0.447} \\
Adenoma high-grade dysp.             & \TablePoint{\underline{0.605}} & \TablePoint{0.684} & \TablePoint{0.387} \\
Stroma                               & \TablePoint{0.604} & \TablePoint{0.581} & \TablePoint{0.528} \\
Adenocarcinoma low-grade             & \TablePoint{0.601} & \TablePoint{0.637} & \TablePoint{0.491} \\
Normal epithelium                    & \TablePoint{0.596} & \TablePoint{0.608} & \TablePoint{0.541} \\
Muscle                               & \TablePoint{0.587} & \TablePoint{0.518} & \TablePoint{0.619} \\
Debris                               & \TablePoint{0.552} & \TablePoint{0.456} & \TablePoint{0.586} \\
Fibrosis                             & \TablePoint{0.525} & \TablePoint{0.246} & \TablePoint{0.846} \\
Lymphocyte                           & \TablePoint{0.525} & \TablePoint{0.402} & \TablePoint{0.600} \\
Inflammation                         & \TablePoint{0.500} & \TablePoint{0.720} & \TablePoint{0.280} \\
Hyperplastic polyp                   & \TablePoint{0.500} & \TablePoint{0.160} & \TablePoint{0.848} \\
Adenoma low-grade dysp.& \TablePoint{0.500} & \TablePoint{0.000} & \TablePoint{1.000} \\
Mucus                   & \TablePoint{0.500} & \TablePoint{0.000} & \TablePoint{1.000} \\
Necrosis               & \TablePoint{0.500} & \TablePoint{0.000} & \TablePoint{1.000} \\
Sessile serrated lesion & \TablePoint{0.500} & \TablePoint{0.000} & \TablePoint{1.000} \\
Vessels                              & \TablePoint{0.466} & \TablePoint{0.370} & \TablePoint{0.599} \\
\bottomrule
\end{tabularx}
\end{table}

\subsection{Single-Model and Ensemble Performance}
\label{sec:results-ensemble}

The MorphContrast model achieved an AUC of $0.642$, balanced
accuracy of $0.594$, sensitivity of $0.635$, and specificity of $0.554$
(Table~\ref{tab:ensemble}). The post hoc ensemble achieved
a pooled OOF AUC of $0.70$ ($95\%$ CI: $0.65$--$0.76$), with balanced accuracy,
sensitivity, and specificity of $0.68$, $0.68$, and $0.68$, respectively.

\begin{table}[!htbp]
\centering
\renewcommand{\baselinestretch}{1}\footnotesize
\footnotesize
\caption{\textbf{Performance of the
MorphContrast model and post hoc ensemble.} Single-model values are
five-fold cross-validation means
$\pm$ standard deviations. Ensemble values are computed
from pooled OOF predictions, with $1{,}000$-sample bootstrap $95\%$ confidence
intervals.}
\label{tab:ensemble}
\setlength{\tabcolsep}{3.5pt}
\newcommand{\TablePoint}[1]{{\fontsize{9}{11}\selectfont\ensuremath{#1}}}
\newcommand{\TableCI}[1]{{\fontsize{6.5}{8}\selectfont\ensuremath{#1}}}
\fontsize{8.5}{10.5}\selectfont
\renewcommand{\arraystretch}{1.08}
\begin{tabularx}{\linewidth}{@{}>{\raggedright\arraybackslash}Xcccc@{}}
\toprule
Model & AUC $\uparrow$ & Bal-Acc $\uparrow$ & Sen $\uparrow$ &
Spec $\uparrow$ \\
\midrule
MorphContrast
& \TablePoint{{\underline{0.642}\pm0.05}}
& \TablePoint{{0.594\pm0.036}}
& \TablePoint{{0.635\pm0.092}}
& \TablePoint{{0.554\pm0.076}} \\
MorphContrast (ensemble)
& \TablePoint{\mathbf{0.70}} \TableCI{(0.65,0.76)}
& \TablePoint{0.68} \TableCI{(0.63,0.73)}
& \TablePoint{0.68} \TableCI{(0.59,0.73)}
& \TablePoint{0.68} \TableCI{(0.62,0.73)} \\
\bottomrule
\end{tabularx}
\end{table}

\subsection{Comparison with MIL and Morphology-Aware Baselines}
\label{sec:results-compare}

Among the conventional MIL methods, TransMIL achieved the highest AUC
($0.67\pm0.05$), followed by CLAM ($0.66\pm0.06$) and ABMIL
($0.63\pm0.06$) (Table~\ref{tab:compare}). TransMIL
yielded $5.25\%$ sensitivity and $99.74\%$ specificity,
while CLAM yielded $25.61\%$ sensitivity and $84.68\%$
specificity. The task-specific model of Nakanishi et al. achieved an AUC of
$0.61\pm0.06$, and RRT-MIL achieved $0.59\pm0.05$.

The morphology-aware baselines achieved AUCs of $0.68\pm0.08$ for
MorphDistill and $0.70\pm0.04$ for PRISM. The MorphContrast model achieved
$0.642\pm0.05$, while the post hoc ensemble achieved a pooled OOF AUC of
$0.70$. For C-index, PRISM achieved $0.672$, followed by the MorphContrast
ensemble ($0.665$), MorphDistill ($0.661$), and TransMIL ($0.647$).

\begin{table}[!htbp]
\centering
\renewcommand{\baselinestretch}{1}\footnotesize
\footnotesize
\caption{\textbf{Comparison with MIL, morphology-aware, and task-specific CRC
prognosis baselines.} Baselines and the single MorphContrast model report mean
$\pm$ standard deviation over folds; the ensemble reports pooled OOF
performance.}
\label{tab:compare}
\setlength{\tabcolsep}{1.5pt}
\newcommand{\TablePoint}[1]{{\fontsize{9}{11}\selectfont\ensuremath{#1}}}
\newcommand{\TableCI}[1]{{\fontsize{6.5}{8}\selectfont\ensuremath{#1}}}
\fontsize{8.5}{10.5}\selectfont
\renewcommand{\arraystretch}{1.08}
\begin{tabularx}{\linewidth}{@{}>{\raggedright\arraybackslash}Xccccc@{}}
\toprule
Model
& AUC $\uparrow$
& Bal-Acc (\%) $\uparrow$
& Sen (\%) $\uparrow$
& Spec (\%) $\uparrow$
& C-index $\uparrow$ \\
\midrule
ABMIL~\cite{ilse2018abmil}
& \TablePoint{0.63\pm0.06} & \TablePoint{58.54\pm4.50} & \TablePoint{51.18\pm9.83} & \TablePoint{65.89\pm4.50} & \TablePoint{0.607} \\
CLAM~\cite{lu2021clam}
& \TablePoint{0.66\pm0.06} & \TablePoint{55.15\pm3.37} & \TablePoint{25.61\pm8.18} & \TablePoint{84.68\pm8.89} & \TablePoint{0.603} \\
TransMIL~\cite{shao2021transmil}
& \TablePoint{0.67\pm0.05} & \TablePoint{52.50\pm4.64} & \TablePoint{5.25\pm9.11} & \TablePoint{99.74\pm0.05} & \TablePoint{0.647} \\
Nakanishi~\cite{nakanishi}
& \TablePoint{0.61\pm0.06} & \TablePoint{58.31\pm5.77} & \TablePoint{53.49\pm11.11} & \TablePoint{63.13\pm2.75} & \TablePoint{0.604} \\
RRT-MIL~\cite{tang2024rrtmil}
& \TablePoint{0.59\pm0.05} & \TablePoint{56.76\pm4.46} & \TablePoint{48.58\pm23.92} & \TablePoint{64.94\pm22.19} & \TablePoint{0.575} \\
MorphDistill~\cite{morphdistill}
& \TablePoint{\underline{0.68}\pm0.08} & \TablePoint{63.41\pm4.70} & \TablePoint{60.24\pm5.38} & \TablePoint{66.57\pm8.81} & \TablePoint{0.661} \\
PRISM~\cite{prism2026}
& \TablePoint{\mathbf{0.70}\pm0.04} & \TablePoint{68.00\pm1.79} & \TablePoint{67.14\pm8.26} & \TablePoint{68.86\pm8.62} & \TablePoint{0.672} \\

MorphContrast 
& \TablePoint{0.642\pm0.05} & \TablePoint{59.45\pm3.56} & \TablePoint{63.53\pm9.23} & \TablePoint{55.37\pm7.59} & \TablePoint{0.616} \\
MorphContrast (ensemble)
& \TablePoint{\mathbf{0.70}} \TableCI{(0.65,0.76)}
& \TablePoint{0.68} \TableCI{(0.63,0.73)}
& \TablePoint{0.68} \TableCI{(0.59,0.73)}
& \TablePoint{0.68} \TableCI{(0.62,0.73)} & \TablePoint{\underline{0.665}} \\
\bottomrule
\end{tabularx}
\end{table}

\subsection{Comparison with Frozen Pathology Foundation Encoders}
\label{sec:results-foundation}

Using the same gated-attention survival head, AUCs for the ten frozen pathology
foundation encoders ranged from $0.51$ to $0.63$
(Table~\ref{tab:foundation}). UNI~v2 achieved the highest AUC
($0.63\pm0.03$), followed by UNI~v1 ($0.62\pm0.05$), H-Optimus-0
($0.61\pm0.04$), and GigaPath ($0.60\pm0.04$). CONCH~v1.5 achieved an AUC
of $0.51\pm0.07$. The MorphContrast model achieved $0.642\pm0.05$
under the same survival-head design. MorphDistill and PRISM, included as
morphology-aware reference models, achieved AUCs of $0.68\pm0.08$ and
$0.70\pm0.04$, respectively; the MorphContrast ensemble
achieved a pooled OOF AUC of $0.70$ ($95\%$ CI: $0.65$--$0.76$).

\begin{table}[!htbp]
\centering
\renewcommand{\baselinestretch}{1}\footnotesize
\footnotesize
\caption{\textbf{Frozen pathology foundation encoders under an identical
survival head.} Each encoder's patch features are aggregated using the same
gated-attention survival head. Values are mean $\pm$ standard deviation over
five folds; the ensemble reports
pooled OOF performance.}
\label{tab:foundation}
\setlength{\tabcolsep}{6pt}
\newcommand{\TablePoint}[1]{{\fontsize{9}{11}\selectfont\ensuremath{#1}}}
\newcommand{\TableCI}[1]{{\fontsize{6.5}{8}\selectfont\ensuremath{#1}}}
\fontsize{8.5}{10.5}\selectfont
\renewcommand{\arraystretch}{1.08}
\begin{tabularx}{\linewidth}{@{}>{\raggedright\arraybackslash}Xcccc@{}}
\toprule
Encoder
& AUC $\uparrow$
& Bal-Acc (\%) $\uparrow$
& Sen (\%) $\uparrow$
& Spec (\%) $\uparrow$ \\
\midrule
CONCH~v1.5~\cite{lu2024conch}
& \TablePoint{0.51\pm0.07} & \TablePoint{50.81\pm3.83} & \TablePoint{41.80\pm15.76} & \TablePoint{59.81\pm16.66} \\
CTransPath~\cite{wang2022ctranspath}
& \TablePoint{0.58\pm0.02} & \TablePoint{50.63\pm1.97} & \TablePoint{21.90\pm39.22} & \TablePoint{79.36\pm37.38} \\
Lunit ViT-S/8~\cite{kang2023lunit}
& \TablePoint{0.58\pm0.04} & \TablePoint{53.22\pm3.14} & \TablePoint{54.24\pm15.52} & \TablePoint{52.21\pm12.69} \\
Phikon~v2~\cite{filiot2024phikon2}
& \TablePoint{0.58\pm0.04} & \TablePoint{58.05\pm5.67} & \TablePoint{59.50\pm15.38} & \TablePoint{56.61\pm11.42} \\
Virchow2~\cite{zimmermann2024virchow2}
& \TablePoint{0.58\pm0.02} & \TablePoint{55.73\pm3.84} & \TablePoint{52.71\pm12.97} & \TablePoint{58.75\pm19.85} \\
Kaiko ViT-B/8~\cite{kaiko2024}
& \TablePoint{0.59\pm0.11} & \TablePoint{58.38\pm8.91} & \TablePoint{46.45\pm30.02} & \TablePoint{70.31\pm13.85} \\
GigaPath~\cite{xu2024gigapath}
& \TablePoint{0.60\pm0.04} & \TablePoint{56.37\pm2.20} & \TablePoint{35.86\pm6.10} & \TablePoint{76.89\pm4.69} \\
H-Optimus-0~\cite{saillard2024hoptimus}
& \TablePoint{0.61\pm0.04} & \TablePoint{56.59\pm4.13} & \TablePoint{36.71\pm19.22} & \TablePoint{76.46\pm18.38} \\
UNI~v1~\cite{chen2024uni}
& \TablePoint{0.62\pm0.05} & \TablePoint{53.58\pm2.44} & \TablePoint{38.14\pm20.97} & \TablePoint{69.02\pm21.79} \\
UNI~v2~\cite{chen2024uni}
& \TablePoint{0.63\pm0.03} & \TablePoint{60.58\pm3.48} & \TablePoint{44.62\pm8.81} & \TablePoint{76.53\pm3.88} \\

MorphDistill~\cite{morphdistill}
& \TablePoint{\underline{0.68}\pm0.08} & \TablePoint{63.41\pm4.70} & \TablePoint{60.24\pm5.38} & \TablePoint{66.57\pm8.81} \\
PRISM~\cite{prism2026}
& \TablePoint{\mathbf{0.70}\pm0.04} & \TablePoint{68.00\pm1.79} & \TablePoint{67.14\pm8.26} & \TablePoint{68.86\pm8.62} \\

MorphContrast 
& \TablePoint{0.64\pm0.05} & \TablePoint{59.45\pm3.56} & \TablePoint{63.53\pm9.23} & \TablePoint{55.37\pm7.59} \\
MorphContrast (ensemble)
& \TablePoint{\mathbf{0.70}} \TableCI{(0.65,0.76)}
& \TablePoint{0.68} \TableCI{(0.63,0.73)}
& \TablePoint{0.68} \TableCI{(0.59,0.73)}
& \TablePoint{0.68} \TableCI{(0.62,0.73)} \\
\bottomrule
\end{tabularx}
\end{table}

\subsection{Comparison with Clinical Variables}
\label{sec:clinical}%
Using the same five patient-disjoint folds ($n=424$; $103$ events),
we compared MorphContrast with $\ell_2$-regularized clinical
logistic-regression models (Supplementary Table~3). Models using all fourteen
variables or nodal burden, T category, and grade only achieved pooled AUCs
of $0.541$ and $0.550$, respectively (Table~\ref{tab:clinical}).
The single MorphContrast model achieved a mean fold AUC of
$0.642\pm0.05$, and the ensemble achieved a pooled OOF AUC of $0.70$.

\begin{table}[!htbp]
\centering
\renewcommand{\baselinestretch}{1}\footnotesize
\caption{\textbf{Clinical-variable models and MorphContrast on Alliance/CALGB~89803 ($n=424$; $103$ events).} Fold AUC reports the mean $\pm$ standard deviation across five test folds. Pooled AUC uses out-of-fold predictions, with DeLong $95\%$ CIs for the clinical models and a $1{,}000$-sample bootstrap $95\%$ CI for the ensemble. Variable coding and imputation are detailed in Supplementary Table~3.}
\label{tab:clinical}
\setlength{\tabcolsep}{3.5pt}
\newcommand{\TablePoint}[1]{{\fontsize{9}{11}\selectfont\ensuremath{#1}}}
\newcommand{\TableCI}[1]{{\fontsize{6.5}{8}\selectfont\ensuremath{#1}}}
\fontsize{8.5}{10.5}\selectfont
\renewcommand{\arraystretch}{1.08}
\begin{tabularx}{\linewidth}{@{}>{\raggedright\arraybackslash}Xccc@{}}
\toprule
Model
& Fold AUC $\uparrow$
& Pooled AUC [95\% CI] $\uparrow$
& C-index $\uparrow$ \\
\midrule
All fourteen variables
& \TablePoint{0.548\pm0.066}
& \TablePoint{0.541} \TableCI{(0.474, 0.608)}
& \TablePoint{0.556} \\

Nodal burden, T category, grade
& \TablePoint{0.554\pm0.052}
& \TablePoint{0.550} \TableCI{(0.487, 0.613)}
& \TablePoint{0.553} \\

MorphContrast
& \TablePoint{0.642\pm0.050}
& ---
& \TablePoint{0.616} \\

MorphContrast (ensemble)
& ---
& \TablePoint{0.70} \TableCI{(0.65, 0.76)}
& \TablePoint{0.665} \\
\bottomrule
\end{tabularx}
\end{table}

\subsection{Risk-Group Stratification}
\label{sec:results-km}

\begin{figure}[!htbp]
\centering

\includegraphics[width=\linewidth]{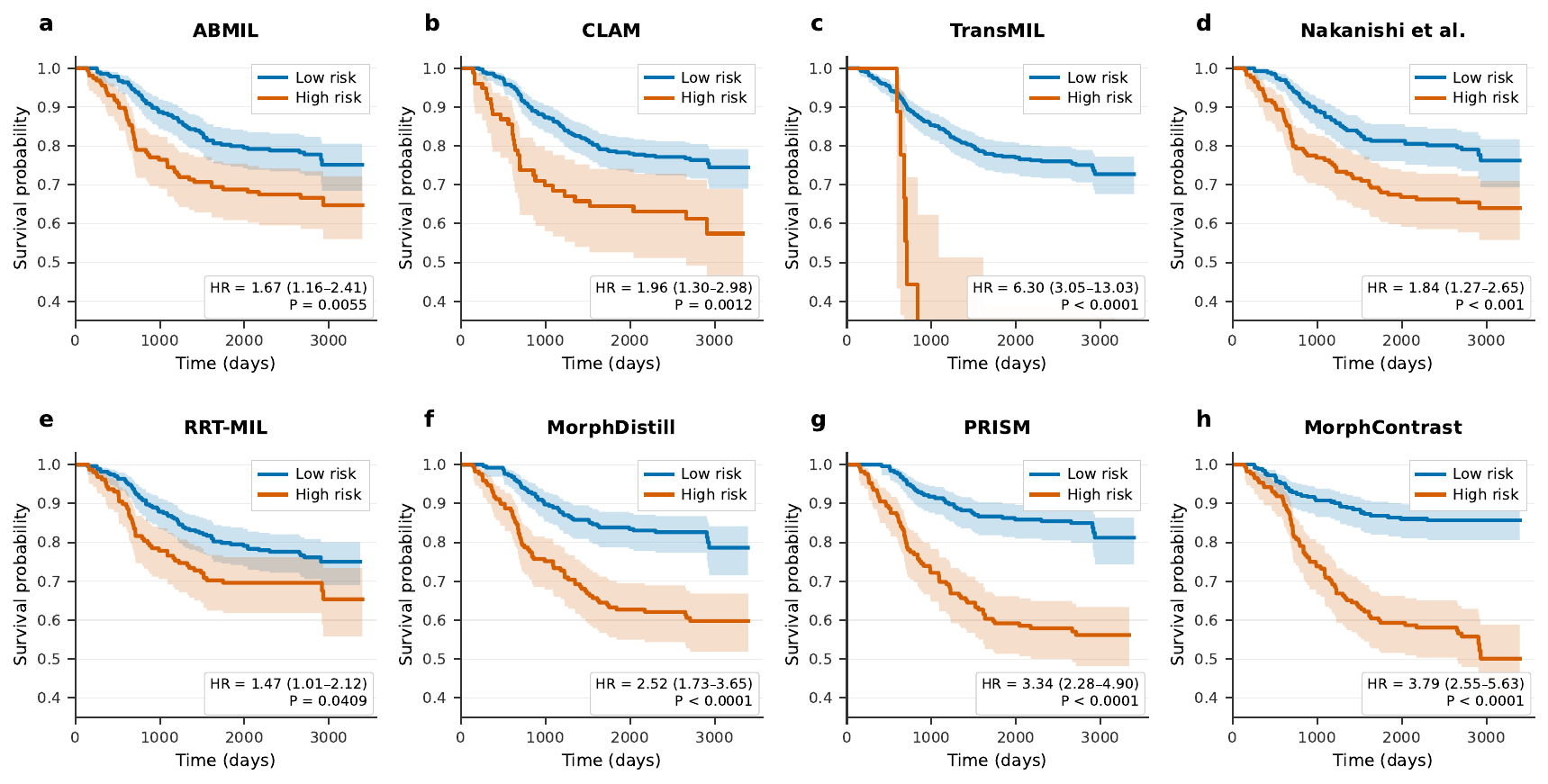}

\caption{\textbf{Kaplan--Meier risk stratification for MorphContrast and
comparison models.} Patients are assigned to low- and high-risk groups at each
model's operating threshold; group sizes are shown in each panel. Hazard ratios
and $95\%$ CIs are obtained from univariate Cox models, and $p$-values from
log-rank tests.}
\label{fig:km}
\end{figure}

\begin{figure}[!htbp]
\centering
\includegraphics[width=\linewidth]{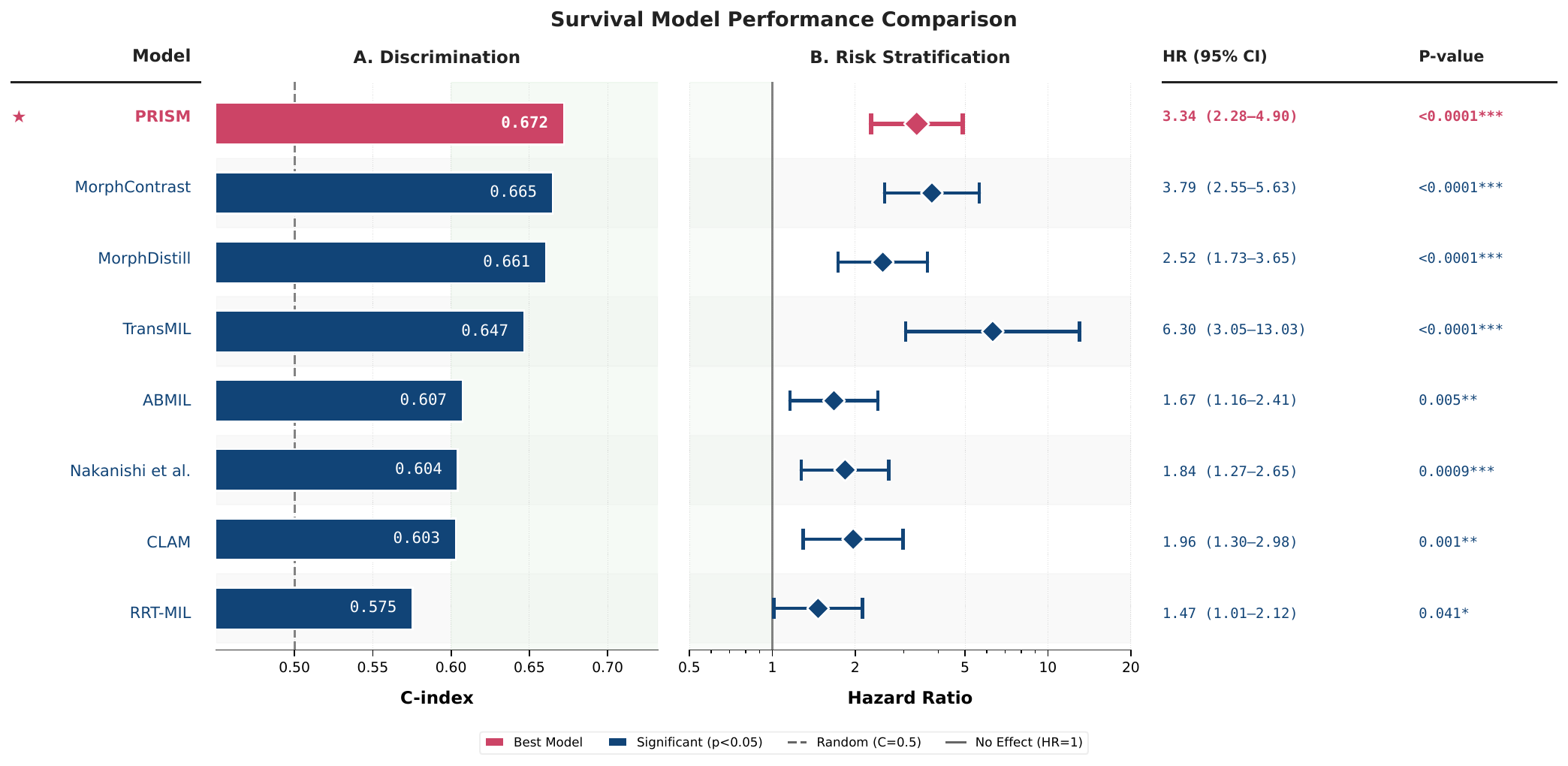}
\caption{\textbf{Discrimination and risk-group separation across MIL and
morphology-aware models.} Left: C-index. Right: univariate Cox hazard ratios
with $95\%$ CIs for high- versus low-risk groups.}
\label{fig:forest}
\end{figure}

The MorphContrast ensemble stratified the Alliance
cohort into low- and high-risk groups, with an HR of $3.79$ ($95\%$ CI:
$2.55$--$5.63$, $p<0.0001$) (Figs.~\ref{fig:km}
and~\ref{fig:forest}). PRISM and MorphDistill
yielded HR estimates of $3.34$ ($95\%$ CI:
$2.28$--$4.90$) and $2.52$ ($95\%$ CI: $1.73$--$3.65$), respectively.
TransMIL yielded an HR of $6.30$ ($95\%$ CI: $3.05$--$13.03$) with nine
patients assigned to the high-risk group; its corresponding operating point was
$5.25\%$ sensitivity and $99.74\%$ specificity.

Among the evaluated frozen foundation encoders, UNI~v2 yielded the largest hazard-ratio point estimate (HR
$2.08$, $95\%$ CI: $1.43$--$3.02$, $p<0.0001$), followed by Phikon~v2
(HR $1.94$, $95\%$ CI: $1.30$--$2.91$) and GigaPath (HR $1.80$, $95\%$
CI: $1.23$--$2.64$) (Supplementary Fig.~1).

\subsection{External Evaluation on the SurGen Cohort}
\label{sec:results-surgen}
We trained all models reported for the SurGen colon cancer cohort exclusively on the Alliance/CALGB~89803 cohort and evaluated them on SurGen in inference mode, without any fine-tuning or adaptation to the external data. This inference-only evaluation stringently tests generalizability under realistic multi-site distribution shift, spanning differences in scanner vendor, resolution, staining, and case-mix composition. On this cohort, FR-MIL achieved the highest
baseline AUC ($0.6766$, $95\%$ CI: $0.597$--$0.754$) and yielded $83.33\%$ sensitivity and $40.27\%$ specificity (Table~\ref{tab:surgen}). TransMIL and MHIM-MIL achieved AUCs of $0.5580$ ($95\%$ CI:
$0.472$--$0.638$) and $0.5585$ ($95\%$ CI: $0.480$--$0.645$),
respectively. The MorphContrast model achieved an AUC of
$0.669$ ($95\%$ CI: $0.58$--$0.75$), with
a balanced accuracy of $63.98\%$, sensitivity of
$57.58\%$, and specificity of $70.39\%$, while the MorphContrast post hoc ensemble achieved an AUC of
$0.678$ ($95\%$ CI: $0.60$--$0.76$), with
a balanced accuracy of $65.80\%$, sensitivity of
$65.15\%$, and specificity of $66.45\%$.

\begin{table}[!htbp]
\centering
\renewcommand{\baselinestretch}{1}\footnotesize
\footnotesize
\caption{\textbf{External validation performance for five-year overall survival
prediction in the SurGen colon cancer cohort.} Models were trained using the Alliance cohort and externally
evaluated on the independent SurGen cohort. The best
AUC point estimate is shown in bold, and the second-best AUC point estimate is
underlined.}
\label{tab:surgen}
\setlength{\tabcolsep}{3.5pt}
\newcommand{\TablePoint}[1]{{\fontsize{9}{11}\selectfont\ensuremath{#1}}}
\newcommand{\TableCI}[1]{{\fontsize{6.5}{8}\selectfont\ensuremath{#1}}}
\fontsize{8.5}{10.5}\selectfont
\renewcommand{\arraystretch}{1.08}
\begin{tabularx}{\linewidth}{@{}>{\raggedright\arraybackslash}Xcccc@{}}
\toprule
Model
& AUC $\uparrow$
& Bal-Acc (\%) $\uparrow$
& Sen. (\%) $\uparrow$
& Spec. (\%) $\uparrow$ \\
\midrule
TransMIL~\cite{shao2021transmil}
& \TablePoint{0.558} \TableCI{(0.47, 0.64)}
& \TablePoint{55.06} \TableCI{(45.1, 57.7)}
& \TablePoint{65.15} \TableCI{(52.2, 76.4)}
& \TablePoint{44.97} \TableCI{(37.6, 53.0)} \\

FR-MIL~\cite{chikontwe2022feature}
& \TablePoint{\underline{0.676}} \TableCI{(0.59, 0.75)}
& \TablePoint{58.80} \TableCI{(46.5, 60.0)}
& \TablePoint{83.33} \TableCI{(73.3, 92.0)}
& \TablePoint{40.27} \TableCI{(32.6, 48.6)} \\

MHIM-MIL~\cite{tang2023mhim}
& \TablePoint{0.558} \TableCI{(0.48, 0.64)}
& \TablePoint{53.61} \TableCI{(47.4, 61.4)}
& \TablePoint{51.52} \TableCI{(40.0, 63.2)}
& \TablePoint{55.70} \TableCI{(47.9, 64.5)} \\

MorphContrast
& \TablePoint{0.669} \TableCI{(0.58, 0.75)}
& \TablePoint{63.98} \TableCI{(60.1, 72.3)}
& \TablePoint{57.58} \TableCI{(30.3, 85.1)}
& \TablePoint{70.39} \TableCI{(44.0, 96.2)} \\

MorphContrast (ensemble)
& \TablePoint{\mathbf{0.678}} \TableCI{(0.60, 0.76)}
& \TablePoint{65.80} \TableCI{(60.7, 73.4)}
& \TablePoint{65.15} \TableCI{(43.9, 85.5)}
& \TablePoint{66.45} \TableCI{(45.4, 87.4)} \\
\bottomrule
\end{tabularx}
\end{table}

\subsection{Ablation of the Contrastive Objectives}
\label{sec}

To assess the individual and joint contributions of the contrastive objectives to five-year survival prediction, we retrained MorphContrast using the same five-fold splits with $\Lintra$ and $\Linter$ enabled individually and jointly, while holding $\Lsurv$ and $\Ldiv$ fixed (Table~\ref{tab:loss_ablation}). The configuration without either contrastive objective reached an AUC of $0.60\pm0.07$ but favored predictions of death within five years, with a sensitivity of $(82.12\pm7.06)\%$ and a specificity of $(22.71\pm7.01)\%$. Either term alone yielded only modest increases in AUC ($0.62\pm0.09$ for $\Lintra$ and $0.61\pm0.08$ for $\Linter$). The full objective achieved the highest mean AUC ($0.64\pm0.05$) and the most balanced operating point among the ablation configurations, with a sensitivity of $(63.52\pm9.23)\%$ and a specificity of $(55.40\pm7.63)\%$. The sensitivity--specificity gap narrowed from $59.41$ to $8.12$ percentage points. The performance gain thus emerges when both contrastive objectives are combined, rather than when either is used alone, and is driven primarily by improved specificity.

\begin{table}[!htbp]
\centering
\renewcommand{\baselinestretch}{1}\footnotesize
\begingroup
\caption{\textbf{Ablation of the intra- and inter-patient contrastive losses
(Alliance/CALGB~89803).} Each configuration adds the indicated contrastive
term to the shared survival ($\mathcal{L}_{\text{surv}}$) and diversity
($\mathcal{L}_{\text{div}}$) objectives under identical five-fold splits;
\ding{51}/\ding{55} mark inclusion. ``Survival only'' omits both contrastive
terms; ``Full'' is complete MorphContrast. Weights are fixed at the established
values ($\lambda_{\text{surv}}{=}1.0$, $\lambda_{\text{div}}{=}0.1$;
$\lambda_{\text{intra}}{=}0.25$, $\lambda_{\text{inter}}{=}0.1$ when enabled).
Values are mean $\pm$ standard deviation\ across folds; in the AUC column the best value is
shown in bold, and the second-best is underlined. AUC is highest for the full
objective, whereas either contrastive term alone is within noise of the
survival-only baseline.}
\label{tab:loss_ablation}
\fontsize{9}{11}\selectfont
\setlength{\tabcolsep}{3pt}
\renewcommand{\arraystretch}{1.1}
\begin{tabularx}{\linewidth}{@{}>{\raggedright\arraybackslash}Xcccccc@{}}
\toprule
& \multicolumn{2}{c}{\textbf{Loss terms}}
& \multicolumn{4}{c}{\textbf{Five-year survival classification}} \\
\cmidrule(lr){2-3}\cmidrule(lr){4-7}
\textbf{Configuration}
& $\mathcal{L}_{\text{intra}}$
& $\mathcal{L}_{\text{inter}}$
& \textbf{AUC}
& \textbf{Bal.\ Acc. (\%)}
& \textbf{Sens. (\%)}
& \textbf{Spec. (\%)} \\
\midrule
Survival only
& \ding{55}
& \ding{55}
& $0.60 \pm 0.07$
& $52.43 \pm 1.32$
& $82.12 \pm 7.06$
& $22.71 \pm 7.01$ \\

$+\,\mathcal{L}_{\text{intra}}$
& \ding{51}
& \ding{55}
& $\underline{0.62} \pm 0.09$
& $53.62 \pm 2.42$
& $72.01 \pm 1.75$
& $35.04 \pm 1.09$ \\

$+\,\mathcal{L}_{\text{inter}}$
& \ding{55}
& \ding{51}
& $0.61 \pm 0.08$
& $56.62 \pm 5.20$
& $80.53 \pm 18.41$
& $32.70 \pm 19.02$ \\

\addlinespace[2pt]
\textbf{Full}
& \ding{51}
& \ding{51}
& $\mathbf{0.64} \pm 0.05$
& $59.43 \pm 3.61$
& $63.52 \pm 9.23$
& $55.40 \pm 7.63$ \\
\bottomrule
\end{tabularx}
\endgroup
\end{table}

\section{Discussion}
\label{sec:discussion}

This study provides three main observations. First, morphology-restricted models
showed modest but distributed prognostic discrimination across tumor and
non-tumor tissue compartments, indicating that
prognostic discrimination was not confined to malignant epithelium. Adipose
tissue and stroma achieved AUCs of $0.622$ and $0.604$,
respectively, compared with $0.601$--$0.605$ for adenocarcinoma
compartments. These results are consistent with
established roles for the invasive front, stromal composition, tumor budding,
and the broader tumor microenvironment in CRC
outcome~\cite{lugli2017itbcc,gao2021tsr,kather2019}. However, these
morphology-specific AUCs should not be interpreted as direct measures of
biological importance. Patch abundance, spatial context, slide sampling, and
correlated clinicopathologic factors were not controlled. The analysis
therefore supports distributed prognostic discrimination
across histologic compartments, not a causal ranking of tissue types.

Second, MorphContrast structures frozen patch features
through two contrastive objectives. Within patients, same-morphology patches
form positive pairs and different-morphology patches form negative pairs. In
the outcome-aware objective, morphology-matched patches from different patients
with concordant outcomes form positive pairs, those with discordant outcomes
form hard negatives, and patches with different morphologies form standard
negatives. On Alliance/CALGB~89803, the model achieved an AUC of
$0.642$, whereas the post hoc ensemble achieved a pooled
OOF AUC of $0.70$ and a risk-group HR of $3.79$.

The MorphContrast model achieved a higher mean AUC than all ten frozen pathology foundation encoders evaluated with the same gated-attention survival head (range, $0.51$--$0.63$), but a lower mean AUC than MorphDistill ($0.68$) and PRISM ($0.70$); however, the post hoc ensemble achieved a pooled OOF AUC of $0.70$, exceeding MorphDistill and matching PRISM.

Third, evaluation on SurGen assessed external performance
in a cohort that differed from Alliance in scanner resolution. Both the MorphContrast model and the post hoc ensemble had higher AUC point estimates
than TransMIL and MHIM-MIL. The $95\%$ confidence interval for the post hoc
ensemble overlapped the interval for FR-MIL, the strongest evaluated baseline.
These external results support cross-cohort generalizability, with
discrimination comparable to the strongest evaluated baseline.

Despite the favorable performance of our method, several limitations warrant consideration. First,
morphology labels are provided by a frozen encoder, so label noise may
propagate directly into the contrastive pairs; ensuring consistently
high-quality positive and negative sets may require further methodological
refinement. Second, the endpoint is a complete-case binary label: patients
censored before five years were excluded rather than censored, which discards
time-to-event information and may introduce selection bias. Finally, the
framework is unimodal; integrating genomic and transcriptomic data, including
microsatellite instability status and consensus molecular subtypes, is a
natural extension~\cite{akbar2026genoglyph}. We intend to address these limitations in future work.

Overall, MorphContrast organizes frozen patch features
through within-patient contrasts defined by morphology and cross-patient
contrasts defined jointly by morphology matching and outcome concordance or
discordance. The prognostic performance observed on Alliance and independently
assessed on SurGen supports the use of morphology labels as structural
supervision for five-year survival prediction from WSIs.

\section{Conclusion}
\label{sec:conclusion}

MorphContrast uses tissue morphology as structural supervision for five-year
survival prediction from CRC H\&E WSIs. The
TCGA-COAD/READ analysis showed modest but distributed prognostic discrimination
across tumor and non-tumor morphologies, motivating morphology- and
outcome-aware contrastive learning. On Alliance/\allowbreak CALGB~89803, the model
achieved an AUC of $0.642$, while the
internally selected post hoc ensemble reached a pooled out-of-fold AUC of $0.70$. On the external
SurGen colon cancer cohort, the single model and ensemble achieved AUCs of $0.669$
and $0.678$, respectively, compared with $0.6766$ for FR-MIL, the strongest
evaluated external baseline. The methodological
contribution of MorphContrast is to use same-patient morphology matches and
mismatches as positive and negative pairs, respectively, and cross-patient,
morphology-matched patches with concordant and discordant outcomes as positive
pairs and hard negatives, respectively. This design converts morphology labels
into outcome-aware supervision for learning prognostic representations from the
complete mixture of tumor and non-tumor tissue.

\section*{Statements and Declarations}
\subsection*{Competing interests}
All authors declare no financial or non-financial competing interests.

\subsection*{Ethics approval and consent to participate}
This study was reviewed and approved by the Institutional Review Board of
The Ohio State University (IRB \#2018C0098). All procedures involving human
data were conducted in accordance with the ethical standards of the
institutional research committee, applicable national
regulations, and the 1964 Declaration of Helsinki and its later amendments. Given the
retrospective nature of the study and because data were archival and
de-identified, the requirement for informed consent to participate was
formally waived by the Institutional Review Board.

\subsection*{Funding}
The project described was supported in part by R01 CA276301
(PIs: Niazi and Chen) from the National Cancer Institute. The project was
also supported in part by Pelotonia under IRP CC13702
(PIs: Niazi, Vilgelm, and Roy), The Ohio State University Department of
Pathology and Comprehensive Cancer Center. The content is solely the
responsibility of the authors and does not necessarily represent the
official views of the National Cancer Institute, National Institutes of
Health, or The Ohio State University.

\subsection*{Author contributions}
H.K. performed data preprocessing, experimental design, validation, and
manuscript writing and editing. M.N.G., A.P., W.L.F., W.C., and
M.K.K.N. assisted with manuscript review and editing.

\subsection*{Acknowledgements}
The authors thank Dr. Fang-Shu Ou and the Alliance Statistics and Data Management Center for data acquisition, curation, and management. The authors also gratefully acknowledge the Ohio Supercomputer Center for providing high-performance computing resources under its contract with The Ohio State University College of Medicine. We thank the Department of Pathology and the Comprehensive Cancer Center at The Ohio State University for their valuable support. The authors also thank Usama Sajjad, a Ph.D. student at The Ohio State University Wexner Medical Center, for providing the baseline results for the Alliance and SurGen cohorts.

\subsection*{Trial registration}
NCT00003835 (CALGB 89803); Registry: CTRP (Clinical Trial Reporting Program);
Registry Identifier: NCI-2012-01844.

\subsection*{Disclaimer}
The data from CALGB 89803 were obtained directly from the Alliance for
Clinical Trials in Oncology, a National Clinical Trials Network cooperative
group; however, all analyses and conclusions in this manuscript are the sole
responsibility of the authors and do not necessarily reflect the opinions or
views of the clinical trial investigators, the NCTN, the NCORP, or the NCI.

\subsection*{Data availability}
The
Alliance/CALGB 89803 whole-slide images and associated clinical outcomes
are available to qualified investigators through the Alliance for Clinical
Trials in Oncology via
\href{mailto:Datasharing@alliancenctn.org}{Datasharing@alliancenctn.org}
upon reasonable request. The remaining cohorts are publicly available: the
TCGA-COAD and TCGA-READ data can be accessed through the GDC Data Portal
(\url{https://portal.gdc.cancer.gov/}), and the SurGen data through the
BioStudies repository
(\url{https://www.ebi.ac.uk/biostudies/bioimages/studies/S-BIAD1285}).

\subsection*{Code availability}
The source code will be publicly available upon acceptance.
Code access may also be requested by emailing the corresponding author at
\href{mailto:Hikmat.khan@osumc.edu?cc=Hikmat.khan179@gmail.com}
{Hikmat.khan@osumc.edu}.

\nocite{chen2026rangersparselygatedmixtureofexpertsadaptive}
\bibliography{references}

\clearpage
\begingroup
\setcounter{figure}{0}
\setcounter{table}{0}
\renewcommand{\figurename}{Supplementary Fig.}
\renewcommand{\tablename}{Supplementary Table}
\renewcommand{\theHfigure}{supplement.\arabic{figure}}
\renewcommand{\theHtable}{supplement.\arabic{table}}
\phantomsection
\pdfbookmark[1]{Supplementary Material}{supplementary-material}
\begin{center}
{\Titlefont Supplementary Material\\[6pt]Morphology-aware deep learning predicts colorectal cancer survival from whole-slide images\par}
\end{center}
\vspace{12pt}
\begin{figure}[!htbp]
\centering
\includegraphics[width=\linewidth]{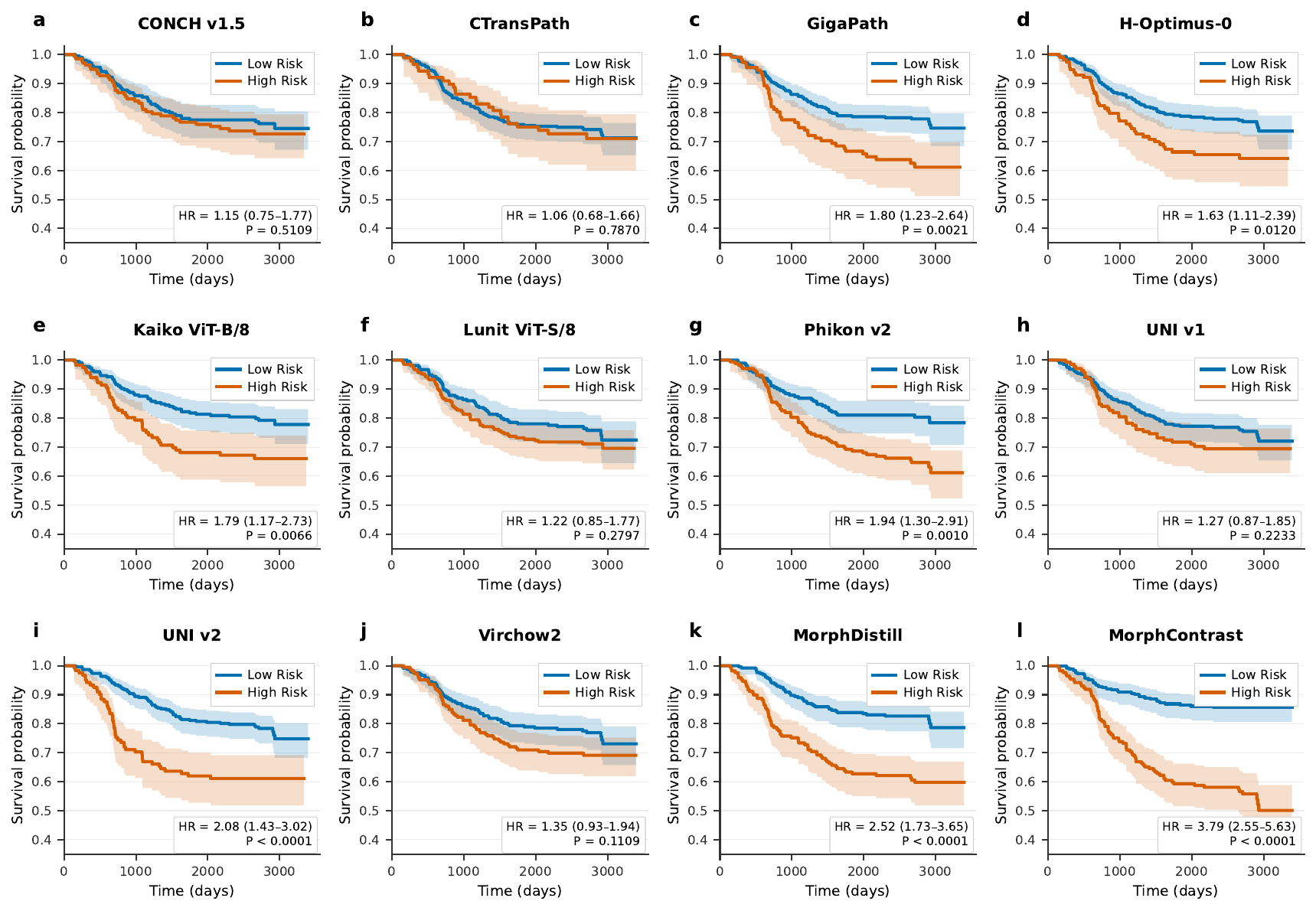}
\caption{\textbf{Kaplan--Meier risk stratification using frozen pathology
foundation encoders and MorphContrast.} Each panel reports low- and high-risk
group sizes, the hazard ratio with $95\%$ CI, and the log-rank $p$-value. The
foundation-model features are aggregated using the same gated-attention survival
head.}
\label{fig:km-foundation}
\end{figure}

\clearpage
\begin{table}[!htbp]
\centering
\caption{\textbf{Clinicopathologic characteristics of the
Alliance/CALGB~89803 cohort.}}
\label{tab:alliance-cohort}
\scriptsize
\setlength{\tabcolsep}{4.5pt}
\renewcommand{\arraystretch}{1.12}
\begin{tabularx}{\linewidth}{
    @{}
    >{\raggedright\arraybackslash}X
    >{\raggedright\arraybackslash}X
    >{\raggedleft\arraybackslash}p{1.8cm}
    @{}
}
\toprule
\multicolumn{3}{c}{\textbf{Alliance cohort characteristics}} \\
\midrule
\textbf{Characteristic} & \textbf{Category} & \textbf{Value} \\
\midrule
Number of slides                    &                 & 424 \\
Number of patients                  &                 & 424 \\
Mean age (years)                    &                 & 60.47 \\
Mean household income median (USD)  &                 & 43,194.59 \\

\addlinespace
\multirow{2}{*}{Race}
    & White & 391 \\
    & Black & 33 \\

\addlinespace
\multirow{2}{*}{Sex}
    & Male & 238 \\
    & Female & 186 \\

\addlinespace
\multirow{2}{*}{Treatment}
    & 5FU/LV & 212 \\
    & CPT-11/5FU/LV & 212 \\

\addlinespace
\multirow{3}{*}{Zubrod performance scale}
    & 0 & 324 \\
    & 1 & 98 \\
    & 2 & 2 \\

\addlinespace
\multirow{7}{*}{Tumor location}
    & Cecum & 101 \\
    & Ascending colon & 64 \\
    & Hepatic flexure & 28 \\
    & Transverse colon & 46 \\
    & Splenic flexure & 19 \\
    & Descending colon & 19 \\
    & Sigmoid colon & 147 \\

\addlinespace
\multirow{4}{*}{Grade}
    & I & 20 \\
    & II & 296 \\
    & III & 108 \\
    & IV & 0 \\

\addlinespace
Stage
    & III & 424 \\

\addlinespace
\multirow{2}{=}{Small blood/lymphatic vessel invasion}
    & No & 286 \\
    & Yes & 138 \\

\addlinespace
\multirow{2}{=}{Extramural vascular invasion}
    & No & 395 \\
    & Yes & 29 \\

\addlinespace
\multirow{2}{*}{Infiltrating border}
    & No & 283 \\
    & Yes & 141 \\

\addlinespace
\multirow{2}{=}{Five-year outcome (analysis endpoint)}
    & Died within 5 years & 103 \\
    & Survived $\geq 5$ years & 321 \\

\bottomrule
\end{tabularx}
\end{table}

\clearpage
\begin{table}[!htbp]
\centering
\small
\caption{\textbf{Hyperparameter search space and fixed training settings for
MorphContrast.} Parameters with multiple candidate values were varied during
hyperparameter search; single-valued parameters were fixed.}
\label{tab:hyperparameters}
\setlength{\tabcolsep}{6pt}
\renewcommand{\arraystretch}{1.05}
\begin{tabular}{lll}
\toprule
\textbf{Parameter} & \textbf{Value(s)} & \textbf{Setting} \\
\midrule
\multicolumn{3}{l}{\textit{Hyperparameter search space}} \\
\addlinespace[2pt]
Feature-encoder hidden dimension
    & $\{768,1024\}$ & Searched \\
Projection dimension
    & $\{128,256\}$ & Searched \\
Attention hidden dimension
    & $\{512,1024,1536\}$ & Searched \\
Initial learning rate
    & $\{10^{-4},10^{-5}\}$ & Searched \\
Decision threshold
    & $\{0.4,0.5\}$ & Searched \\
Intra-patient loss weight $\lambda_a$
    & $\{0.1,0.25\}$ & Initial search \\
Inter-patient loss weight $\lambda_e$
    & $\{0.1,0.25\}$ & Initial search \\
Diversity loss weight $\lambda_d$
    & $\{0.1,0.25\}$ & Searched \\
Survival loss
    & Weighted CE, focal loss & Searched \\
\midrule
\multicolumn{3}{l}{\textit{Fixed training settings}} \\
\addlinespace[2pt]
Optimizer
    & AdamW & Fixed \\
Learning-rate schedule
    & Cosine annealing & Fixed \\
Weight decay
    & $10^{-5}$ & Fixed \\
Batch size
    & $4$ & Fixed \\
Dropout
    & $0.1$ & Fixed \\
Survival loss weight $\lambda_s$
    & $1.0$ & Fixed \\
Temperature $\tau$
    & $0.1$ & Fixed \\
Maximum contrastive patches
    & $512$ & Fixed \\
Maximum epochs
    & $150$ & Fixed \\
Early-stopping patience
    & $15$ epochs & Fixed \\
\bottomrule
\end{tabular}
\end{table}

\clearpage
\begin{table}[!htbp]
\centering
\small
\caption{\textbf{Clinicopathological variables used in the clinical models.} The fourteen variables were obtained from Alliance/CALGB~89803 trial case-report forms ($n=424$; $103$ events). Both full and reduced baselines use $\ell_2$-regularized logistic regression to predict death within five years, evaluated using the same five patient-disjoint folds as MorphContrast. Missing values were imputed with medians estimated from the training patients in each fold. The reduced model includes positive lymph-node count, T category, and grade only. T category combines trial depth-of-invasion codes $4$ and $5$ as T4. FL denotes fluorouracil and leucovorin; IFL additionally includes irinotecan. No molecular markers were included.}
\label{tab:clinical-variables}
\setlength{\tabcolsep}{6pt}
\renewcommand{\arraystretch}{1.12}
\begin{tabularx}{\linewidth}{@{}>{\raggedright\arraybackslash}X>{\raggedright\arraybackslash}X@{}}
\toprule
\textbf{Variable} & \textbf{Model coding} \\
\midrule
Age at enrolment & years, continuous \\
Sex & male vs female \\
Histological grade & III (poorly differentiated) vs I--II \\
T category (depth of invasion) & T4 vs T1--T3 \\
Positive lymph nodes & number, continuous \\
Small blood/lymphatic vessel invasion & present vs absent \\
Extramural vascular invasion & present vs absent \\
Perineural invasion & present vs absent \\
Infiltrating border & present vs absent \\
Lymphocytic response & present vs absent \\
Obstruction & present vs absent \\
Perforation & present vs absent \\
Zubrod performance status & $\geq 1$ vs 0 \\
Treatment arm & IFL vs FL \\
\bottomrule
\end{tabularx}
\end{table}

\clearpage
\endgroup
\end{document}